# Single and Cross-Dimensional Feature Detection and Description: An Evaluation

Odysseas Kechagias-Stamatis [1,2*], Nabil Aouf [1], Mark A. Richardson [2]

[1] Centre for Electronic Warfare Information and Cyber, Cranfield University Defence and Security, SN6 8LA, Shrivenham, UK
[2] Department of Electrical and Electronic Engineering, City University of London, EC1V 0HB, London, UK
*Odysseas.Kechagias-Stamatis@city.ac.uk

**Abstract:** Three-dimensional local feature detection and description techniques are widely used for object registration and recognition applications. Although several evaluations of 3D local feature detection and description methods have already been published, these are constrained in a single dimensional scheme, i.e. either 3D or 2D methods that are applied onto multiple projections of the 3D data. However, cross-dimensional (mixed 2D and 3D) feature detection and description has yet to be investigated. Here, we evaluated the performance of both single and cross-dimensional feature detection and description methods on several 3D datasets and demonstrated the superiority of cross-dimensional over single-dimensional schemes.

## 1. Introduction

Local features in 3D data have been widely investigated to improve the distinctiveness and robustness of local feature (keypoint) detection and description methods. Given the importance of these methods for 3D data registration and classification applications, it is necessary to evaluate keypoint detectors and feature descriptors [1]. Most such evaluations have been presented in the context of reports comparing current methods to newly proposed techniques, although some studies dedicated to the evaluation of 3D keypoint detectors or feature descriptors have also been published [2–4]. However, such evaluations have been limited to a single domain, with 3D methods applied directly to 3D data [2–6], or 2D methods applied to multiple 2D projections of 3D data [7–11].

Examples of the direct 3D approach include the evaluation of several 3D keypoint detectors by comparing the robustness of each technique to rotation, scaling and translation [5], an evaluation focusing on the optimum combination of 3D keypoint detection and feature description [6], and a complete and thorough evaluation of 3D keypoint detectors, with further limited evaluation carried out of selected 3D descriptors [2]. The most comprehensive studies of 3D feature descriptors reported thus far also included work involving a selection of 3D keypoint detectors [5, 6]. These latter reports represent the most comprehensive evaluations of 3D keypoint detection [2] and description methods [5, 6] published thus far.

In an example of the indirect approach (2D schemes applied to 3D data, where the 3D data are presented in a 2D range image form), state-of-the-art 2D descriptors have been evaluated via several transformations of the initial 2D range image, including maximum curvature, mean curvature and shape index [7]. The authors found that Scale Invariant Feature Transform (SIFT) [8] achieved the best performance in terms of facial recognition whereas Fast Retina Keypoint (FREAK) [9] achieved the best trade-off between performance and speed. The evaluation of 2D methods on projections of 3D data in point cloud form has also been attempted, but only in the context of comparing current methods to newly proposed techniques [10, 11].

The performance of 2D and 3D schemes has been evaluated independently without cross-dimensional keypoint detection and feature description or the direct comparison of pure 3D and 2D schemes. Cross-dimensional evaluation has not been attempted yet and refers to challenging both 2D and 3D local keypoint detection and feature description methods against 3D data using a cross-dimensional approach. The comparison of 3D and 2D schemes has been superficially addressed in the context of comparing a proposed technique against 3D methods [12, 13]. Driven by the absence of such comparisons, we therefore evaluated both single (3D and 2D) and cross-dimensional keypoint detection and feature description on several 3D datasets varying in content and complexity. The aim of this work was to identify potential cross-modality combinations that exploit the advantages of both 2D and 3D methods in terms of robustness and computational efficiency, with performance and computational requirements given equal priority. It should be noted that despite single-modality keypoint detection and feature description on pure 3D and 2D data not originating from projections has already been presented in [2, 4, 14, 15], to make a direct comparison between single and cross-modality comparison feasible, it is necessary to re-evaluate these keypoint detection and feature description methods on the same dataset used in this paper.

The remainder of the article is organized as follows: Section 2 presents the 2D and 3D keypoint detectors and feature descriptors that were evaluated. Section 3 presents the experimental setup and Section 4 evaluates the 2D and 3D techniques in single and cross-dimensional schemes. Our conclusions are presented in Section 5.

## 2. 2D / 3D Keypoint Detection and Feature Description Methods

### 2.1. Keypoint Detectors

Keypoint detectors analyse the structure around a vertex or a pixel depending on the data domain (3D or 2D,

respectively) and classify as keypoints the vertices/pixels that fulfil some specific criteria that are dependent on the detector itself. Ideally, keypoints are prominent among their surroundings, have unique features, and can be redetected even if the object to which they belong is distorted or corrupted.

*2.1.1 2D detectors:*

**Harris**: Harris is a fixed scale corner detector [16], which relies on an autocorrelation function that captures the intensity variations of an image *I* in a neighbourhood window *Q* centred at pixel *p(x,y)*:

$$E(x, y) = \sum_Q w(u,v)\left[I(u+x, v+y) - I(u,v)\right]^2 \quad (1)$$

where *(x, y)* are the pixel coordinates in *I* and *w(u, v)* is the window patch at position *(u, v)*. Using Taylor's approximation, Harris rearranges Eq. (1) as follows:

$$E(x, y) = \begin{bmatrix} x & y \end{bmatrix} M \begin{bmatrix} x \\ y \end{bmatrix} \quad (2)$$

$$M = \begin{bmatrix} \sum_Q I_u^2 & \sum_Q I_u I_v \\ \sum_Q I_u I_v & \sum_Q I_v^2 \end{bmatrix} \quad (3)$$

where $I_u$, $I_v$ represent the spatial gradients of the image.

The shape of *Q* is classified based on the eigenvalues $\lambda_1$ and $\lambda_2$ of *M*. Specifically, if both values are small, *E* also has a small value and *Q* has an approximately constant intensity. If both are large, *E* has a sharp peak indicating that *Q* includes a corner, and if $\lambda_1 > \lambda_2$ then *Q* includes an edge. To measure the corner or edge quality, Harris introduced metric *R*:

$$R(x, y) = \det(M) - k \cdot tr(M) = \lambda_1 \lambda_2 - k(\lambda_1 + \lambda_2) \quad (4)$$

where $k \in [0.04,..,0.15]$.

**Good Features To Track (GFTT)**: Shi and Tomasi [17] extended the robustness of the Harris corner detector by proposing that *Q* encloses a corner if $\min(\lambda_1, \lambda_2) > \lambda$, where λ is a predefined threshold. GFTT, like Harris, is a fixed scale detector.

**Difference of Gaussians (DoG)**: Lowe [8] proposed the SIFT keypoint detection and description scheme. For the keypoint detection part, Lowe extended the work of Lindeberg [18] to detect local extrema in image *I* utilizing a DoG scheme rather than a scale-normalized Laplacian of Gaussian as originally proposed. This modification aims to reduce the overall processing burden during keypoint detection.

For the DoG scheme, a pyramid of images is created to achieve scale invariance by convolving *I* with Gaussian kernels at various scales. The output of two sequential convolutions is subtracted creating a new set of images, i.e. DoG images, in which pixels are classified as candidate keypoints. Then the pixel value of each candidate keypoint is compared with its eight neighbours in the same scale, the nine pixels one scale above and the nine pixels one scale below. If the pixel value of a candidate keypoint has the highest value within its neighbourhood then it is labelled as a keypoint. The latter comparison is the popular *non-maxima suppression* process. Finally, the keypoint detection stage ends with a refinement process to discard keypoints that have a low contrast and that lie on edges. The former are discarded by applying a texture threshold, whereas the latter are discarded by identifying Harris keypoints [16]. DoG is an adaptive scale keypoint detector.

**Fast Hessian (FH)**: A processing-efficient alternative to DoG is the FH detector used as the keypoint detection part of the popular Speeded-UP Robust Features (SURF) algorithm [19]. In order to avoid convolution with second-order derivatives, this technique approximates the Gaussian kernels with their discretized version (i.e. box filters) that are computed with a constant time cost by utilizing the integral image concept [20]. Like the DoG detector, candidate features are obtained after a 3 × 3 × 3 neighbourhood non-maximum suppression process. Finally, candidate keypoints with a response *R* exceeding a pre-defined threshold are preserved while the rest are discarded:

$$R(x, y, \sigma) = D_{xx}(\sigma) D_{yy}(\sigma) - (0.9 D xy(\sigma))^2 \quad (5)$$

where *Dxx(σ), Dyy(σ)* and *Dxy(σ)* are the outputs after convolving the corresponding box filters of standard deviation σ with image *I*. FH is an adaptive scale keypoint detector.

**Features from Accelerated Segment Test (FAST)**: FAST [21] detects keypoints in an image *I* by placing around the pixel of interest *p* a circle that has a circumference of 16 pixels. If $I_p$ is the pixel intensity at pixel *p* and *thresh* a pre-defined threshold, then *p* is labelled as a keypoint if *N-contiguous* pixels in the circle are brighter than $I_p$+*thresh* or darker than $I_p$−*thresh*. FAST is a fixed scale keypoint detector.

**Binary Robust Invariant Scalable Keypoints (BRISK)**: The BRISK technique [22] involves both a keypoint detection and a description scheme. For the former, it uses the FAST [21] keypoint detector in 9-16 mask configuration, i.e. placing around the pixel of interest *p* a circle that has a circumference of 16 pixels, and considering the intensity of nine contiguous pixels within that circle. In BRISK, the FAST technique is combined with maxima suppression applied in a scale-space fashion using the FAST score as a measure of saliency. However, in contrast to DoG and SURF, keypoints are sought within a continuous scale-space by involving not only the true octaves but also virtual intra-octave levels.

**KAZE**: KAZE [23] is similar to SURF in that it relies on the response of a scale-normalized determinant of the Hessian at multiple scale levels, but it involves a non-linear scale-space rather than the linear scale-space used in SURF. KAZE is an adaptive scale keypoint detector.

*2.1.2 3D detectors:*

**Intrinsic Shape Signatures (ISS)**: ISS [24] measures the saliency of a point *p(x,y,z)* based on the eigenvalue decomposition of the scatter matrix Σ(*p*) of the *N* vertices within the support region (neighbourhood) *V* of *p*:

$$\Sigma(p) = \frac{1}{N} \sum_{q \in V} (q - \mu_p)(q - \mu_p)^T \quad (6)$$

$$\mu_p = \frac{1}{N} \sum_{q \in V} q \quad (7)$$

ISS suggests that vertices fulfilling Eq. (8) are labelled as candidate keypoints:

$$\frac{\lambda_2(p)}{\lambda_1(p)} < threshold_1 \wedge \frac{\lambda_3(p)}{\lambda_2(p)} < threshold_2 \quad (8)$$

where the $\lambda_1, \lambda_2, \lambda_3$ are the eigenvalues of $\Sigma(p)$ in order of decreasing magnitude. Finally, candidate keypoints with the smallest eigenvalues and large variation along each principal direction are labelled as ISS keypoints. ISS is a fixed scale keypoint detector.

**KeyPoint Quality (KPQ)**: KPQ is a keypoint detector that ranks candidate keypoints based on a quality metric [25]. Specifically, $V$ is aligned to the canonical reference frame given by the principal directions, and then non-distinctive vertices are discarded by thresholding the ratio between the maximum lengths along the first two principal axes. The remaining vertices are labelled as candidate keypoints, which are then evaluated for their saliency:

$$\rho(p) \doteq \frac{1000}{N^2} \sum_{q \in V} |K(p)| + \max_{q \in V}(100 K(p)) + \min_{q \in V}(100 K(p)) + \max_{q \in V}(10 k_1) + \min_{q \in V}(10 k_2) \quad (9)$$

where $K$ is the Gaussian curvature and $k_1$, $k_2$ are the principal curvatures. Vertices with a $\rho(p)$ value exceeding a threshold and fulfilling certain constraints are labelled KPQ keypoints. These constraints are (i) the minimum Euclidean distance between two KPQ keypoints is greater than a certain threshold; and (ii) that within a support radius only one KPQ can exist. Sensitivity to noise and sampling is reduced by estimating $k_1$ and $k_2$ over smoothed and resampled surfaces that are properly aligned to the original point cloud. KPQ is a fixed scale keypoint detector.

**Harris 3D**: The Point Cloud Library (PCL) [26] community provides a 3D variant of the classic 2D Harris [16]. Although 2D and 3D Harris are conceptually similar, the modification required for extension to a 3D keypoint detector involves substituting the image gradients in the covariance matrix of Eq. (3) with the normal vector of the support region $V$ centred on vertex $p(x,y,z)$ of the point cloud. Harris 3D is a fixed scale keypoint detector.

**Local Surface Patches (LSP)**: LSP [12] uses the Shape Index (ShI) metric [27] to measure the saliency of vertex $p(x,y,z)$. Vertices $p$ that fulfil the following constraint are considered as candidate LSP keypoints:

$$ShI(p) \geq (1+a)\mu_{ShI(p)} \vee ShI(p) \leq (1-\beta)\mu_{ShI(p)} \quad (10)$$

where $\mu_{SI(p)}$ is the average SI of the support region $V$ and $\alpha$, $\beta$ are user-defined thresholds. Candidate LSP keypoints then undergo a non-maxima suppression process and the remaining vertices are classified as LSP keypoints. LSP is a fixed scale keypoint detector.

**Heat Kernel Signature (HKS)**: HKS is a saliency metric based on the restriction of the heat kernel to the temporal domain that is computed on the mesh $M$ of the point cloud [28]. Vertex $p(x,y,z)$ is defined as an HKS keypoint if its saliency $k_{t'}$ at time interval $t'$ fulfils the following constraint:

$$k_{t'}(p,p) > k_{t'}(q,q) \quad (11)$$

where $q$ is a vertex belonging to a two-ring neighbourhood of $p$ and $k_{t'}(p,q)$ is a function that represents the amount of heat transferred from vertex $p$ to $q$ in time $t'$ given a unit heat source at vertex $p$. Thus, $k_{t'}(p,q)$ is governed by the heat equation:

$$\Delta_M u(x,t) = -\frac{\partial u(x,t)}{\partial t} \quad (12)$$

where $\Delta_M u(x,t)$ is the Laplace-Beltrami operator defined on manifold $M$. HKS is a fixed scale keypoint detector.

**Laplace-Beltrami Scale Space (LBSS)**: Unnikrishnan and Hebert [29] classify a vertex $p(x,y,z)$ as a keypoint if its scale-space saliency $\rho(p,t)$ exceeds a certain threshold:

$$\rho(p,t) = \frac{2\|p - A(p,t)\|}{t} e^{-\frac{2\|p-A(p,t)\|}{t}} \quad (13)$$

$$A(p,t) = p + \frac{t^2}{2}\Delta_M p \quad (14)$$

where $\Delta_M$ is the Laplace-Beltrami operator. In simpler terms, $\rho(p,t)$ can be considered as a displacement of $p$ along its normal that is proportional to the mean curvature. LBSS is an adaptive scale keypoint detector and scale-space is implemented by increasing the size of the support region $V$.

**MeshDoG:** MeshDoG [30] is a similar solution to LBSS but scale-space is created using the DoG concept [8]. MeshDoG is applied on a transformed representation of the point cloud, where for the context of this paper we use the mean curvature [2]. The scale-space saliency $\rho(p,t)$ of a vertex $p(x,y,z)$ is defined as:

$$\rho(p,t) = C_H^{(t)}(p) - C_H^{(t-1)}(p) \quad (15)$$

$$C_H^{(t)} = C_H^{(t-1)} * G(\sigma) \quad (16)$$

where $C_H^{(t)}$ is the $t$-th convolution of the mean curvature map $C_H$ with the Gaussian kernel of zero mean and σ standard deviation. MeshDoG is an adaptive scale keypoint detector.

**Salient Points (SP)**: SP [31] is similar to MeshDoG [30] but is directly applied to the vertex coordinates rather than a transformed representation of the point cloud. SP is an adaptive scale keypoint detector.

**KPQ-AS**: This is an extension of the KPQ technique [25] that facilitates adaptive-scale keypoint detection. Scale-space is created by increasing the support region $V$ and scale selection is achieved by performing non-maxima suppression.

### 2.2. Local Feature Descriptors

Local feature description techniques describe local patches around a point of interest by encoding the properties of the local patch. Ideally, feature descriptors describe each keypoint in a unique manner and are robust to nuisance factors such as resolution variation and noise.

*2.2.1 2D descriptors:*

**SIFT:** Lowe [8] describes a keypoint detection method but also suggests a feature description technique. The latter initially assigns to each keypoint one or multiple orientations that are based on the local gradient information. The magnitude and direction of the gradient form an orientation histogram with 36 bins based on the neighbourhood of the keypoint. The histogram is then weighted by a Gaussian kernel that is placed around the keypoint and the peak of the histogram corresponds to the orientation of the keypoint. In the event this histogram has peaks of at least 80% of the main peak, then additional descriptions of the same keypoint are created that share the same scale but have different orientations.

The scale and orientation linked to each keypoint form a local coordinate frame. Specifically, the descriptor is computed using the gradient magnitude and orientations in a $16 \times 16$ window around the keypoint (rotated according to orientation). These are stacked in 8-bin histograms formed in $4 \times 4$ sub-regions and are weighted by a Gaussian window.

**SURF**: SURF [19] initially performs an orientation assignment by computing Gaussian-weighted Haar wavelet responses over a circular region with a radius six times the scale where the keypoint is detected. Once an orientation is assigned, the description process involves a square region ($20 \times$ scale) centred on the keypoint and oriented accordingly. This region is further divided into 4×4 sub-regions and then vertical and horizontal Haar-wavelet responses are computed, which are weighted with a Gaussian kernel. This process is performed at fixed sample points and is summed up in each sub-region. Finally, the polarity of intensity changes is also calculated by summing the absolute values of the horizontal and vertical responses. SURF features of opposing polarity are not matched.

**BRISK**: The BRISK method [22] encodes keypoints using a handcrafted sampling pattern comprising concentric circular patches centred at a keypoint. Aliasing effects during sampling are avoided by applying local Gaussian smoothing on the patch to be described, with a standard deviation proportional to the distance between the circle centre and the keypoint.

There are two types of sampling pairs (short and long pairs) that depend on the distance between them. The long pairs have a distance greater than threshold $d_{min}$, and are used to compute the local gradient (of the patch) that defines the orientation of the feature. The short pairs with a distance less than threshold $d_{max}$ are then rotated accordingly to achieve rotation invariance and are used to compute the binary BRISK descriptor via intensity tests.

**FREAK:** FREAK [9] is a biologically inspired binary keypoint descriptor that applies a series of intensity tests on a patch that encloses the keypoint. FREAK and BRISK share the same sampling pattern and use the same mechanism to estimate the keypoint orientation. However, FREAK is influenced by the human retinal system and exploits a circular sampling grid with sampling points that are denser near the centre and become exponentially less dense further away from the centre. The advantage of this concept is that the test pairs naturally form a coarse-to-fine approach. Feature matching is accelerated by comparing the coarse part of the descriptor and if these exceed a threshold then the fine part is tested.

**KAZE:** The keypoint description part of KAZE [23] is similar to SURF but is properly adapted to facilitate a non-linear scale-space framework.

*2.2.2 3D descriptors:*

The 3D local feature description techniques comprise a support volume $V$ that in centred on a keypoint $p(x,y,z)$ by encoding the geometric properties and the underlying structure of $V$ [32]. Their major advantages include robust feature description for partially visible objects [33] and lower susceptibility to illumination variation and pose changes [34, 35]. The 3D descriptors evaluated herein are described below. However, because we attributed equal importance to performance and processing efficiency, we did not evaluate 3D Shape Context (3DSC) [36] and its extension the Unique Shape Context (USC) [37] due to their high computational burden.

**Histogram of Distances (HoD)/ HoD-Short (HoD-S):** HoD [38] is a robust and processing-efficient 3D descriptor that calculates the probability mass density of the normalized point-pair $L_2$-norm distance distributions within $V$. $L_2$-norm distances are encoded in a coarse and a fine manner by using different bin sizes during distance quantization. Finally, the two types of encodings are concatenated in a single descriptor. This dual encoding scheme enhances feature-matching performance in the presence of noise and subsampling perturbations. HoD does not require a local reference frame (LRF) or axis (LRA) and adapts the description radius on the target point cloud resolution rather than the template, which is the norm for a 3D descriptor. HoD-S [39, 40] is a compact version of HoD that exploits only on the coarse part of HoD.

**Signatures of Histograms of Orientations (SHOT):** SHOT [41] divides the support volume $V$ into a number of sub-volumes along the azimuth, the elevation and the radius. For each sub-volume, a 1D histogram is computed based on the normal variation between the keypoint $p(x,y,z)$ (including its surrounding vertices) and the vertices that lie in each sub-volume.

**Fast Point Feature Histograms (FPFH):** FPFH [42] establishes on $V$ a *Darboux* LRF. Then for each point belonging to $V$, FPFH encodes the angular relationship between the keypoint $p(x,y,z)$ and its neighbours as provided by the LRF. Finally, this angular relationship is transformed into a histogram.

**Rotational Projection Statistics (RoPS):** RoPS [43] establishes on $V$ a LRF, then $V$ is rotated around every axis of the LRF and is projected on each of the coordinate planes. Finally, each projection undergoes a statistical analysis based on low-order moments and entropy, which are converted into a 1D histogram.

**Tri-Spin Images (TriSI):** TriSI [44] is an extension of the popular 3D descriptor Spin Images (SI) [45]. For the latter, given a support volume $V$ centred at point $p(x,y,z)$, a LRA is aligned with the normal vector of the vertices within $V$, a 2D array accumulator with user-defined dimensions is placed on the LRA, and the SI descriptor is generated by accumulating the neighbouring points into each bin of the 2D array as the array *spins* around the LRA. TriSI uses the same technique as SI but substitutes the LRA with an LRF and calculates a SI value for each axis of the LRF. Finally, the three SI values are concatenated to from a TriSI descriptor.

## 3. Length Experimental Setup

### 3.1. Datasets

We evaluated the effectiveness and the robustness of each keypoint detector and feature descriptor by employing two classes of trials, namely registration and object recognition. The former was based on the Oakland dataset [46], whereas the latter was based on the Laser Scanner dataset [34], the Kinect dataset [47] and the SpaceTime dataset [41].

*3.1.1 Oakland dataset:* This dataset comprises 18 point cloud scenes of the Oakland University campus captured using a light detection and ranging (LIDAR) device. For our point cloud registration scenario, we exploit two consecutive scenes that have some overlap. Then one of the two scenes was randomly rotated (pitch, roll and yaw) by up to 180° and simultaneously translated in the *X, Y* and *Z* directions by up to 10 m.

*3.1.2 Laser Scanner dataset:* This is the most cited dataset in the 3D computer vision literature. It comprises five model point clouds and 50 scene point clouds of high quality. Each model comprises a full 3D point cloud, whereas the scenes are 2.5D point clouds (i.e. viewing-dependent point clouds based on a specific vantage point). Scenes also contain clutter objects and the target is occluded.

*3.1.3 Kinect dataset:* The Kinect dataset comprises six models and 16 scenes acquired by a Microsoft Kinect sensor. Given the sensing device, the point cloud quality is low, and the models within scenes are occluded and mixed with clutter objects. In contrast to the Laser Scanner dataset, the models and scenes in the Kinect dataset share the same dimensionality (2.5D).

*3.1.4 SpaceTime dataset:* The SpaceTime dataset [41] was created by using the SpaceTime Stereo technique and comprises eight models and 15 scenes. Given the use of this technique, the point clouds are of medium quality. Each scene encloses the target object which is cluttered and occluded.

### 3.2. Evaluation

Given a model and a target point cloud, the first part of our evaluation involved challenging the 3D keypoint detection methods against the 2D methods. For the former, we applied the 3D keypoint detectors presented in Section 2 to both the Model $M$ and the Target $T$. As previously reported [2], we avoided the influence of border vertices on the keypoint detection process by discarding border keypoints. Then, given the known homography between $M$ and $T$, we calculated a number of performance metrics for each keypoint detector (Section 3.3).

The 3D techniques were applied directly to the point cloud data, whereas for the 2D techniques we initially projected each point cloud onto the main planes of the *XYZ* global reference frame that was fitted to the point cloud during acquisition. Then on each projection, we applied the 2D keypoint detectors presented in Section 2. Finally, we back-projected the detected 2D keypoints of each projection to the initial point cloud and calculated the performance metrics used for the 3D keypoint detectors. During the 3D to multi-2D projection, we properly quantized the coordinates of each vertex to remap the 3D floating-point vertex coordinates $p(x,y,z)$ into pixel coordinates $p_Q(x,y,z)$:

$$p_Q(x,y,z) = \lfloor q_f \cdot p(x,y,z) \rfloor \quad (17)$$

where $q_f$ is the quantization factor and $\lfloor \cdot \rfloor$ the bottom-round process. The overall keypoint evaluation pipeline is presented in Fig. 1 (a).

The second part of our evaluation compared single and cross-dimensional keypoint detection and feature description. Specifically, we evaluated the performance of 3D keypoint detection and description, 2D keypoint detection and description, 3D keypoint detection with 2D description, and 2D keypoint detection with 3D description. We assessed only the top-performing 2D and 3D keypoint detectors based on the results of the first part of our evaluation. As described for the first stage, we applied each 2D descriptor to the projected planes of the *XYZ* global reference frame that was fitted to the point cloud during acquisition. During 2D feature matching, we cross-matched all features from every plane of the model and the target point clouds and created a list of corresponding pixels, which were back-projected into the original 3D domain. The performance metrics for each keypoint detector and feature descriptor combination (Section 3.3) were used for both single and cross-dimensional keypoint detection. Fig. 1 (b) shows the feature description architecture for all single and cross-dimensional keypoint detection and feature matching combinations.

### 3.3. Comparison Metrics

*3.3.1 Absolute / Relative repeatability:* Repeatability is the most important metric for a keypoint detector because it defines its ability to find the same keypoints on different instances of a given 3D point cloud or 2D image. For the 2D and 3D detectors evaluated in this study, we extracted a keypoint $k_M$ from the model $M$ (either a 3D point cloud or a 2D image projection depending on the evaluation) and transformed it into $k_{MT}$ according to the homography, i.e. rotation $R$ and translation $T$, between the model and the scene. A keypoint $k_M$ is repeatable if the Euclidean distance of $k_{MT}$ from its nearest keypoint $k_S$ that is extracted from the scene $S$ is less than a threshold $\varepsilon$.

$$\|Rk_M + T - k_S\| = \|k_{MT} - k_S\| < \varepsilon \quad (18)$$

The absolute repeatability (AR) and the relative repeatability (RR) [4, 48] are defined as:

$$AR = C^+ \quad (19)$$

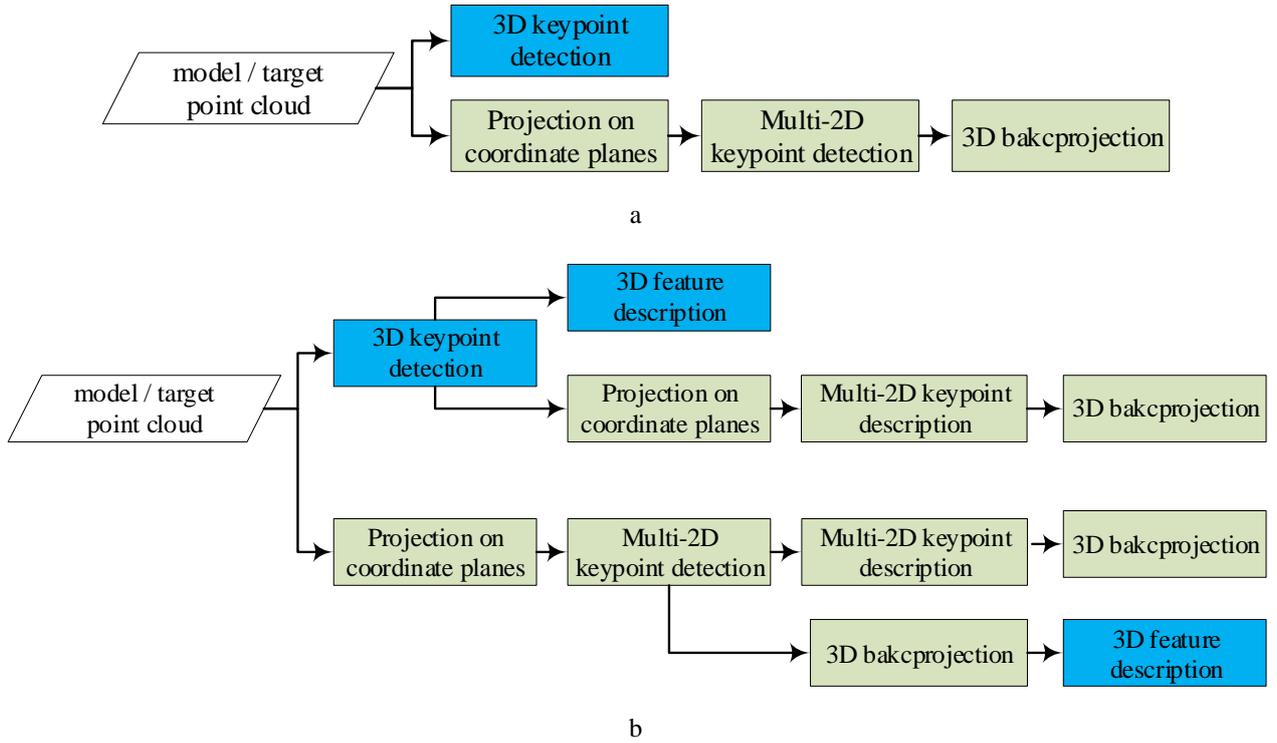

***Fig. 1*** *(a) The 2D/3D keypoint detection pipeline. (b) The 2D/3D keypoint detection and description pipeline (green = 2D process, blue = 3D process).*

$$RR = \frac{C^+}{C} \quad (20)$$

where $C^+$ is the number of keypoints that fulfil Eq. (18) and $C$ the number of detected keypoints in the model scene. Model keypoints $k_M$ were only considered if they were present in the scene. We therefore checked whether a real vertex existed within a small neighbourhood of the fictitious $k_{MT}$ created by the known model-scene homography. If this was true, the real vertex closest to the fictitious $k_{MT}$ was linked with $k_M$. This neighbourhood is defined by a sphere centred at $k_{MT}$ with a radius of $10 \times Tr$, with $Tr$ the average Target point cloud resolution.

In contrast to previous studies [2, 4, 45], we set a larger radius in order to achieve a common neighbourhood size for all tests and also to compensate for transformation errors that occur when the 3D point cloud vertices are projected onto multiple 2D images and are then back-projected to a 3D point cloud.

*3.3.2 Area Under Curve (AUC):* The AUC metric is a single value that indicates the overall performance of the descriptor. Here, we calculated the AUC based on the 1-Precision–Recall (PR) curve [4]. Given a scene feature $f_S$ that encodes the keypoint $k_S$, a list of model features and the model-to-scene ground truth homography, $f_S$ is matched against all model features to find the closest. If the Euclidean distance of the keypoints that have matched features is less than a threshold $\mu$ then the match is considered as a True Positive (TP) otherwise as a False Positive (FP). Features that are incorrectly not matched are labelled as a False Negative (FN). Hence, *1-Precision* and *Recall* are defined as:

$$1 - \Pr ecision = 1 - \frac{TP}{TP + FP} \quad (21)$$

$$\operatorname{Re} call = \frac{TP}{TP + FN} \quad (22)$$

The PR curve is obtained by varying threshold $\mu \in [0,1]$ and matching exploits the Fast Library for Approximate Nearest Neighbours (FLANN) [49].

*3.3.3 Compactness:* This metric relates the descriptive power to the cardinality of a description vector. This is important because the length of the feature vector has a great impact on the memory footprint and computational requirements during the feature matching stage. As previously reported [4], we define *compactness* as:

$$compactness = \frac{Average\ AUC}{Descriptor\ cardinality} \quad (23)$$

*3.4. Implementation*

All trials were performed in MATLAB on an Intel i7 with 16 GB of RAM. Keypoint detectors and descriptors were implemented in either MATLAB or C++/PCL using a MEX wrapper. The tuned parameters of each detector (Table 1) and descriptor (Table 2) were used to maximize performance. The un-tuned parameters were fixed either to those proposed by the original authors or to their PCL implementation [6, 38, 50]. For the tuning process, we used the *Oakland* dataset and confirmed that SHOT has a stable description performance regardless of the description radius, whereas TriSI, FPFH and RoPS gain peak performance and then drop [4]. For the scenarios we evaluated, this peak performance was identified at a radius of $20 \times Mr$, with $Mr$ representing the average Model point cloud resolution. For HoD and HoD-S, optimal performance was achieved at $20 \times Tr$.

**Table 1** Keypoint Detectors Evaluated

| Dimension | Descriptor | Implementation platform | Tuned parameters |
|---|---|---|---|
| 3D | ISS | C++ (MEX wrapper) | - |
| 3D | SP | MATLAB | - |
| 3D | Harris 3D | MATLAB | - |
| 3D | KPQ | MATLAB | - |
| 3D | Uniform | C++ (MEX wrapper) | Grid size of 5 x point cloud resolution |
| 2D | GFTT | C++ (MEX wrapper) | Min. corner quality $10^{-3}$ / Gaussian filter size 3x3 |
| 2D | FAST16-Adaptive scale | C++ (MEX wrapper) | Min. corner quality $10^{-3}$ / Min. intensity contrast $10^{-3}$ / Octaves 4 |
| 2D | FAST6-Fixed scale | C++ (MEX wrapper) | Min. corner quality $10^{-3}$ / Min. intensity contrast $10^{-3}$ |
| 2D | DoG | MATLAB | 8 scale levels |
| 2D | FAST-9 | C++ (MEX wrapper) | Intensity threshold 9 |
| 2D | Harris 2D | C++ (MEX wrapper) | Min. corner quality $10^{-3}$ / Gaussian filter size 3x3 |
| 2D | KAZE | C++ (MEX wrapper) | 6 scale levels / 6 octaves |
| 2D | FH-9 | C++ (MEX wrapper) | 6 scale levels / blob threshold $10^{-5}$ |

**Table 2** Feature Descriptors Evaluated

| Dimension | Descriptor | Descriptor Length | Implementation platform | Tuned parameters |
|---|---|---|---|---|
| 3D | SHOT | 352 | C++ (MEX wrapper) | - |
| 3D | HoD | 240 | MATLAB | - |
| 3D | HoD-S | 40 | MATLAB | - |
| 3D | FPFH | 33 | C++ (MEX wrapper) | - |
| 3D | RoPS | 135 | MATLAB | - |
| 3D | TriSI | 675 | MATLAB | - |
| 2D | FREAK | 64 | C++ (MEX wrapper) | - |
| 2D | SURF | 64 | C++ (MEX wrapper) | - |
| 2D | BRISK | 64 | C++ (MEX wrapper) | - |
| 2D | KAZE | 64 | C++ (MEX wrapper) | - |
| 2D | SIFT | 128 | C++ (MEX wrapper) | 8 scale levels |

## 4. Experimental Results and Discussion

### 4.1. Evaluation of Keypoint Detectors

*4.1.1 Oakland dataset:* One important factor affecting the performance of the 2D keypoint detection methods is the quantization factor $q_f$ used during the 3D to multi-2D remapping process applied to the point cloud. As shown in Fig. 2 (a), the RR increased with $q_f$ for all methods with the exception of DoG, which showed a stable but poor performance. This is because $q_f$ defines the amount of detail preserved on the 2D image projections after the remapping process, with higher $q_f$ values corresponding to a higher resolution. Fig. 2 (b) shows the corresponding AR achieved by each method, revealing that RR and AR have a similar relationship with $q_f$. The low RR performance of DoG reflects the extremely low AR. Due to the log scale of the AR plot, zero AR is omitted and thus AR plots can be interrupted.

The selection of $q_f$ has also a direct impact on the number of detected keypoints, the physical size of the 2D projections and ultimately on the overall processing time required to apply the 2D keypoint detection methods.

Given that we regarded performance and computational efficiency as equally important, we set $q_f = 10$ for the remaining trials. The processing burden for $q_f = 10$ is 57 times lower than $q_f = 100$, but most of the keypoint detection methods still perform well (Fig. 2(a)). Table 3 shows the processing time needed for various $q_f$ values and the process acceleration relative to $q_f = 10$. Fig. 3 shows the processing time needed by each 2D keypoint detector (for $q_f = 10$) and the corresponding time for the 3D keypoint detection methods evaluated herein ($q_f$ is not applicable in the 3D methods). Fig. 3 shows that although the computational time of the 2D methods includes the 3D to multi-2D remapping, keypoint detection process for all three planes, and keypoint back-projection to the original 3D domain, the computational burden is much lower than that of almost all 3D descriptors. Fast9 achieved the lowest processing time, followed by GFTT, Harris and FH. The highest computational burden was associated with KPQ and KPQ-AS.

**Table 3** Overall Processing Time for Various $q_f$ Values

| $q_f$ | 100 | 50 | 20 | 10 | 5 |
|---|---|---|---|---|---|
| avg. time (s) | 19.12 | 6.14 | 1.26 | 0.33 | 0.31 |
| gain factor | 57 | 18 | 4 | 1 | 1 |

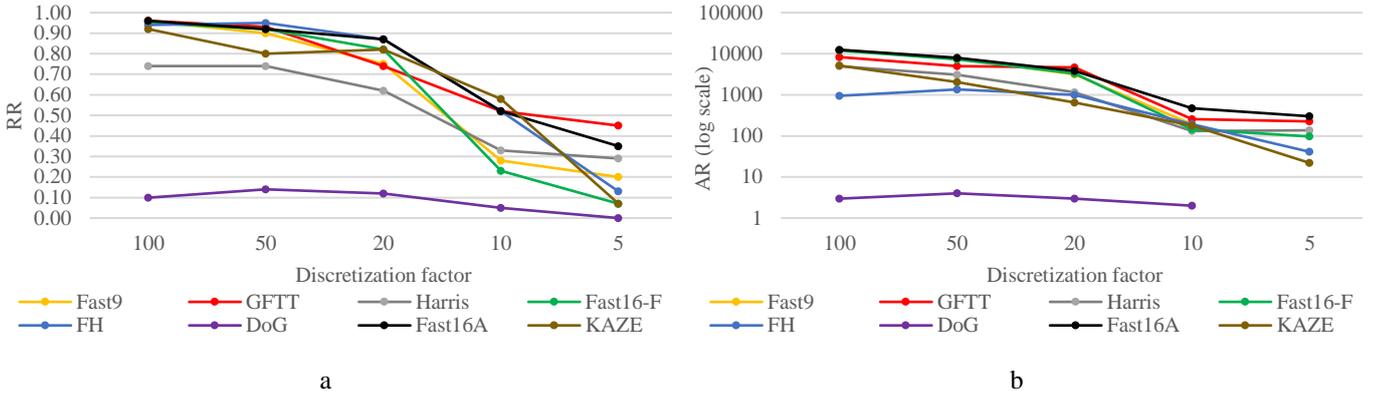

***Fig. 2** Impact of discretization factor on 2D keypoint detection on the Oakland dataset. (a) RR. (b) AR.*

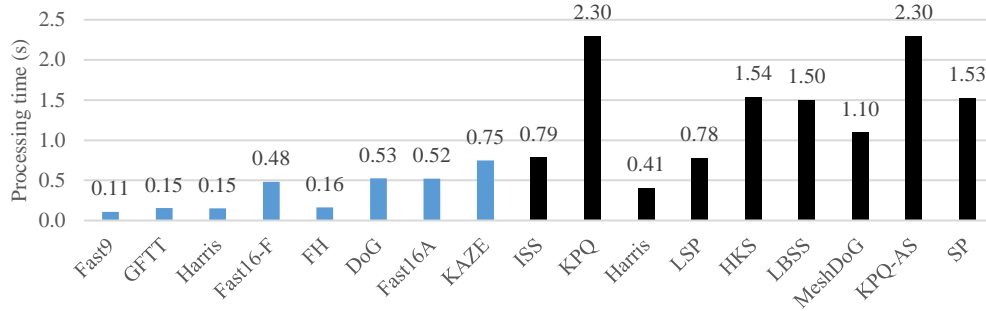

***Fig. 3** Processing time of 2D (blue) and 3D (black) keypoint detectors ( $q_f = 10$ for the 2D methods).*

Subsequent trials on the Oakland dataset evaluated the robustness of the 2D and 3D keypoint detection methods challenged by variable resolution, Gaussian noise and Shot noise. These nuisance factors are added to one of the two scene segments and simultaneously the same segment was also randomly rotated up to 180° in pitch, roll and yaw and translated up to 10 meters in the *X, Y* and *Z* directions, creating a highly complex and challenging scenario that exceeds the typical difficulty of current computer vision scenarios. This complexity was introduced to investigate the limits of the keypoint detection methods and the single and cross-dimensional 2D/3D keypoint detection and description methods described in Section 4.2.

In the nuisance-free setting, the 2D keypoint detection methods achieved an average RR of 38% compared to 22% for the 3D methods, indicating that the 2D methods are more robust to resolution variation (Fig. 4). This is mainly due to the coordinate remapping process in Eq. (17), which transforms the floating-point vertex coordinates into pixel coordinates. Even when the resolution was reduced to one eighth of the original value, most of the 2D methods, namely Fast9, GFTT, Harris, Fast16-F (fixed scale) and Fast16-A (adaptive scale), were still able to achieve appealing RR and AR scores. Interestingly, DoG performed less well than anticipated, but this was due to the extremely small number of keypoints it provides. In contrast, the 3D methods were much more vulnerable to resolution variation even when the resolution was reduced to only half its original value.

Next we investigated the robustness of each method to various Gaussian noise levels with zero mean and standard deviation $\sigma = \{0.1Mr, 0.3Mr, 0.5Mr\}$ [6, 38]. Fig. 5 (a,b) clearly shows that the 2D keypoint detectors were only marginally affected regardless of the noise level, with KAZE, GFTT, Fast16-A and FH demonstrating a highly appealing and stable performance. This is because the low quantization value $q_f = 10$ during the coordinate remapping process of Eq. (17) quantizes the noisy vertex coordinates in the same pixel coordinates as seen in the noise-free case. Unlike the 2D methods, the 3D methods were strongly affected even by low Gaussian noise levels (Fig. 5 (c,d)).

Finally, we evaluated the robustness of each method to various Shot noise levels modelled with a Poisson process where $\lambda = \{0.1Mr, 0.3Mr, 0.5Mr\}$. Fig. 6 shows that the 2D descriptors were only marginally affected, retaining their high RR and AR values. Their appealing performance is yet again due to the quantization process of Eq. (17) and the small $q_f$ value. In contrast, the 3D descriptors were strongly influenced by even low levels of Shot noise.

Regarding the overall performance of the 2D and 3D keypoint detectors on the Oakland dataset, it is evident that the majority of the 2D techniques outperform the 3D ones both in terms of processing efficiency and robustness to resolution and nose variation. In the remaining challenges using alternative datasets, the 2D and 3D keypoint detection methods were tested against the standard dataset alone, without resolution or noise variation.

*4.1.2 Laser Scanner dataset:* When tested against the Laser Scanner dataset, the best RR performance was achieved by the 3D keypoint detector ISS with the 2D detector GFTT following closely behind (Fig. 7 (a)). The 2D and 3D methods achieved average RR values of 21% and 20%, respectively, and in both cases the AR values provided on average a similar number of keypoints (Fig. 7 (b)).

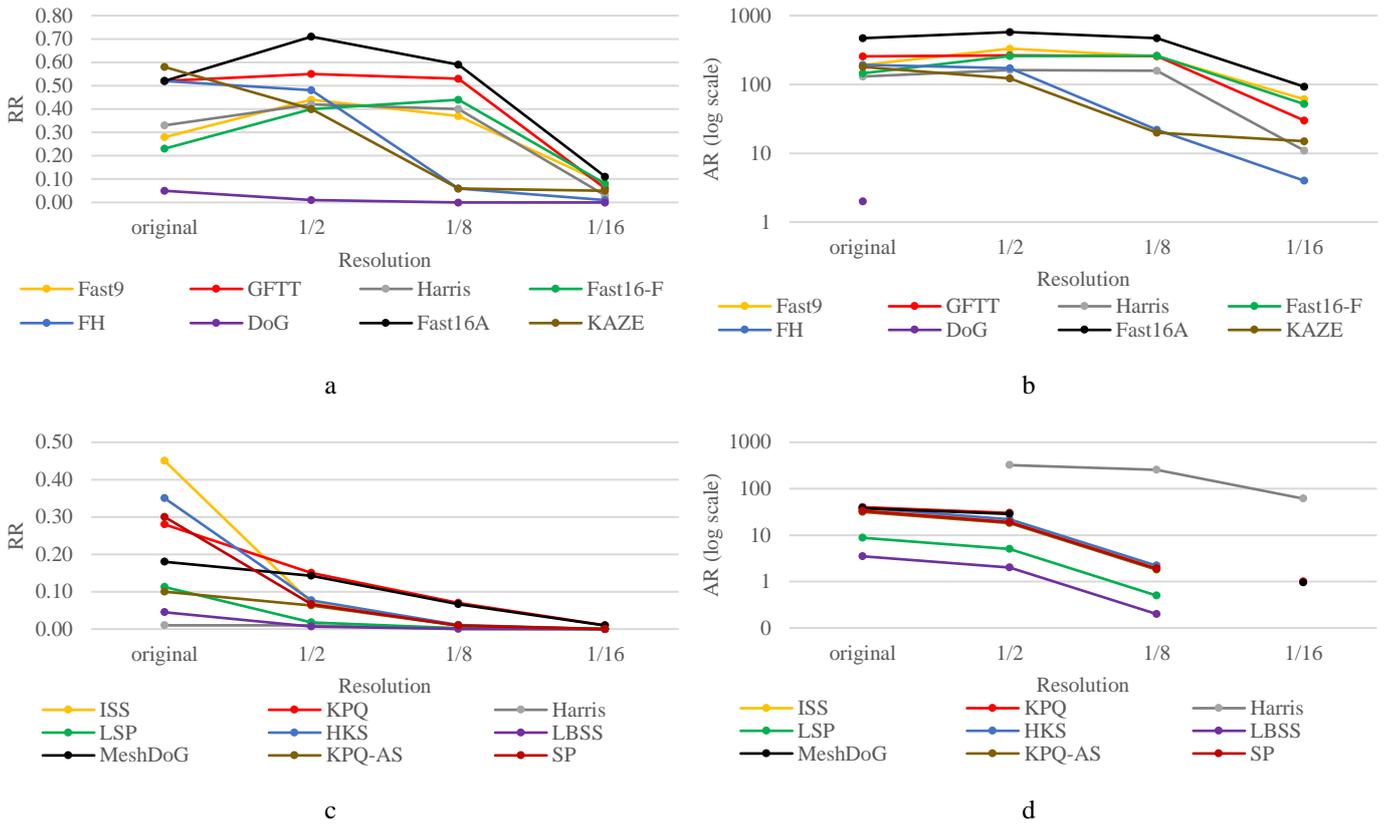

***Fig.4*** *Evaluating detector performance on the Oakland dataset under various resolution levels.*
*(a) 2D techniques RR. (b) 2D techniques AR. (c) 3D techniques RR. (d) 3D techniques AR.*

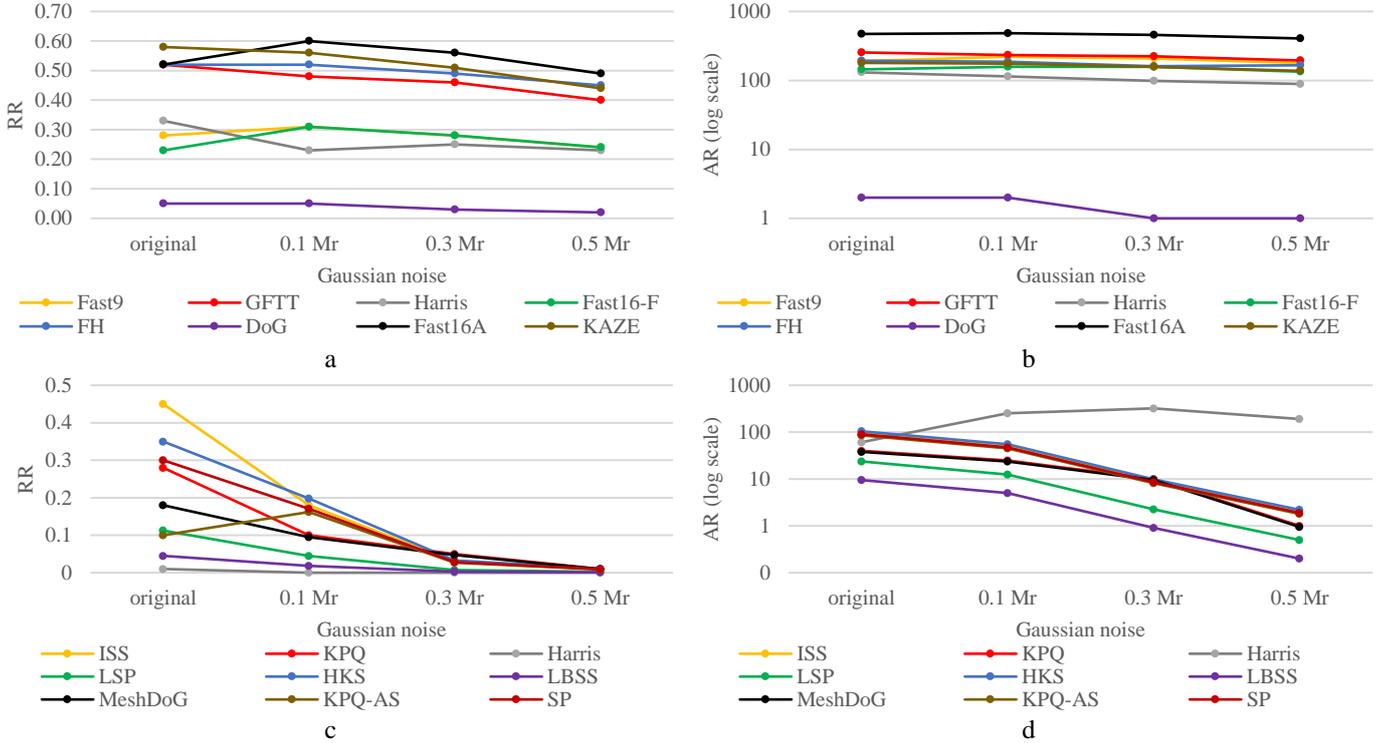

***Fig.5*** *Evaluating detector performance on the Oakland dataset under various Gaussian noise levels.*
*(a) 2D techniques RR. (b) 2D techniques AR. (c) 3D techniques RR. (d) 3D techniques AR.*

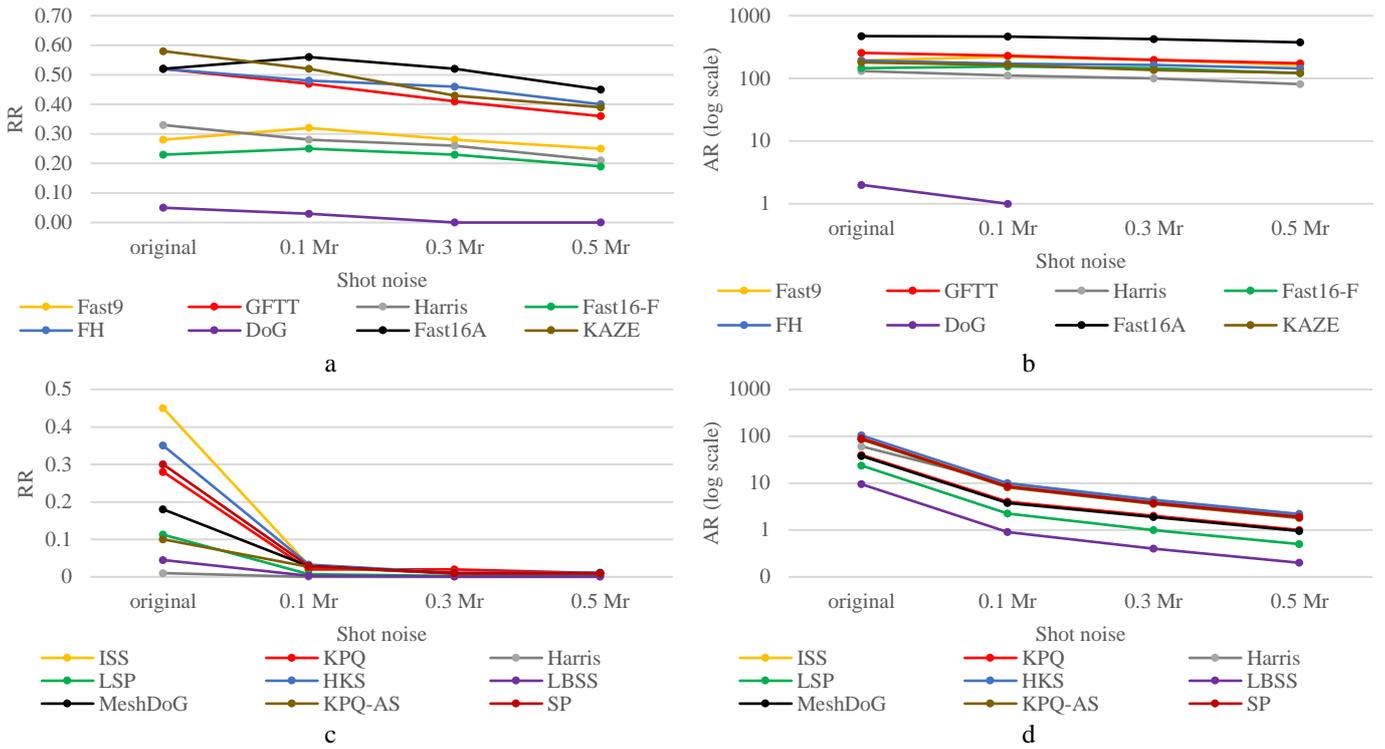

*Fig.6 Evaluating detector performance on the Oakland dataset under various Shot noise levels.*
*(a) 2D techniques RR. (b) 2D techniques AR. (c) 3D techniques RR. (d) 3D techniques AR.*

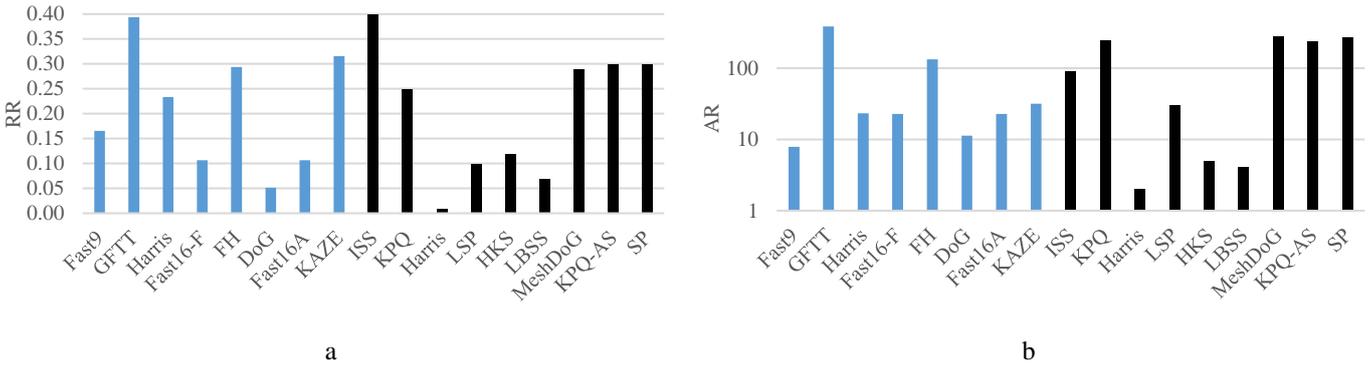

*Fig. 7 Evaluating 2D and 3D keypoint detectors on the Laser Scanner dataset. (a) RR. (b) AR.*

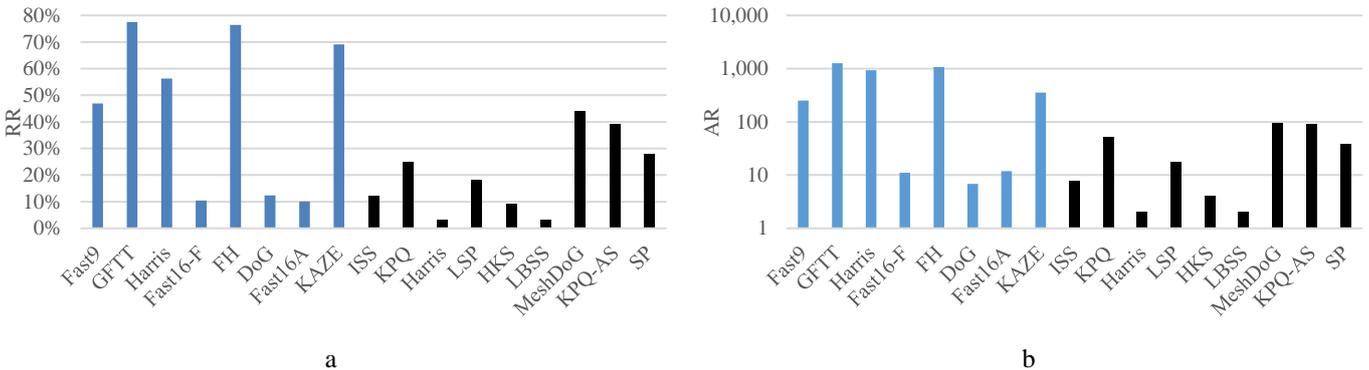

*Fig. 8 Evaluating 2D and 3D keypoint detectors on the Kinect dataset. (a) RR. (b) AR.*

*4.1.3 Kinect dataset:* When tested against the Kinect dataset, most of the 2D keypoint detectors (GFTT, FH, KAZE, Harris and Fast9) achieved a better RR performance than the corresponding 3D methods, with GFTT and FH exceeding 75% RR (Fig. 8). The average RR of the 2D methods (45%) was far superior to the average RR of the 3D methods (22%).

*4.1.4 SpaceTime dataset:* The 2D methods achieved higher RR values than the 3D methods when tested against the SpaceTime dataset, with Fast16-A performing best. The average RR of the 2D methods was 47%, compared to 28% for the 3D methods (Fig. 9).

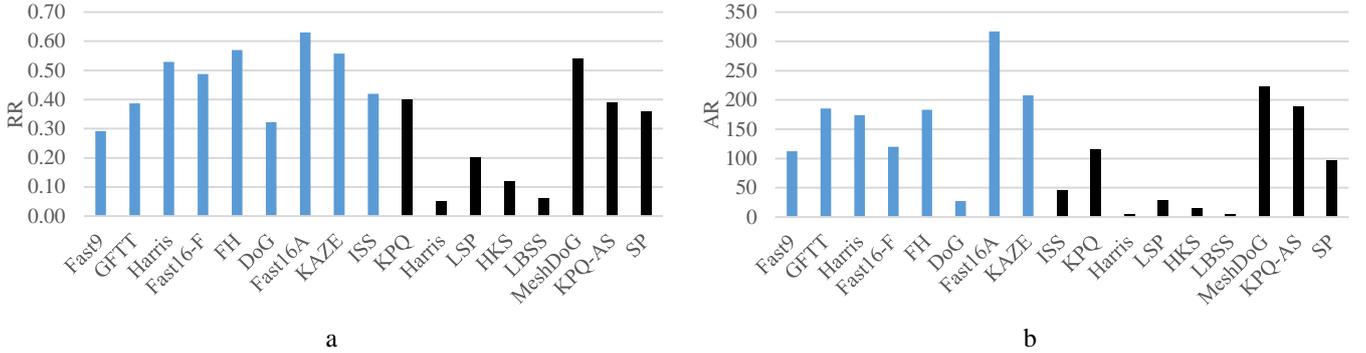

*Fig. 9 Evaluating 2D and 3D keypoint detectors on the SpaceTime dataset. (a) RR (b) AR.*

**Table 4** Computational Time for the OAKLAND Dataset

|  |  |  | 2D descriptors |  |  |  |  | 3D descriptors |  |  |  |  |  |
|---|---|---|---|---|---|---|---|---|---|---|---|---|---|
|  |  |  | FREAK | SURF | BRISK | KAZE | SIFT | HoD-S | HoD | SHOT | FPFH | RoPS | TriSI |
| Keypoint detectors | 2D | GFTT | 0.23 | 0.11 | 0.92 | 0.14 | 0.26 | 2.64 | 4.05 | 8.20 | 54.78 | 26.29 | 44.92 |
|  |  | Fast16-A | 0.16 | 0.05 | 0.80 | 0.07 | 0.26 | 0.83 | 1.17 | 5.19 | 26.51 | 7.68 | 29.60 |
|  |  | FH | 0.16 | 0.05 | 0.80 | 0.07 | 0.05 | 0.65 | 0.93 | 4.82 | 34.34 | 5.94 | 29.61 |
|  | 3D | ISS | 0.29 | 0.20 | 0.94 | 0.29 | 0.26 | 1.84 | 2.74 | 6.54 | 59.46 | 21.39 | 46.35 |
|  |  | Uniform | 0.20 | 0.11 | 0.84 | 0.15 | 0.27 | 0.98 | 1.40 | 5.19 | 53.23 | 9.43 | 35.15 |

*4.1.5 Discussion:* From the keypoint detection trials it is evident that the 2D methods are overall more appealing than the 3D methods for several reasons:

    a. In the nuisance-free *Oakland* and *Kinect* datasets, the RR values of the 2D methods were double those of the 3D methods. For the *Laser Scanner* dataset, both the 2D and 3D methods achieved a similar performance.

    b. The 2D methods were superior in terms of robustness to nuisances (resolution variation, Gaussian and Shot noise). This was mainly due to the low quantization value $q_f = 10$ during the coordinate remapping process of Eq. (17). In contrast, due to the challenging complexity of the *Oakland* scenario, the 3D keypoint detection techniques achieved very low RR values even at the lowest nuisance levels.

    c. The 2D methods were four times faster to execute than the 3D methods, despite the former requiring a multi-staged process that includes 3D to multi-2D remapping, keypoint detection on all three planes and keypoint back-projection to the original 3D domain.

### 4.2. Evaluation of Feature Descriptors

Next, we conducted single and cross-dimensional evaluations of the 2D and 3D keypoint detection and description methods on the datasets described in Section 4.1. The trials comprised 2D-2D, 2D-3D, 3D-2D and 3D-3D schemes, where the first and second numbers indicate the dimensionality of the keypoint detector and feature descriptor, respectively. To improve clarity, only the GFTT, Fast16-A and FH keypoint detector methods were used for the 2D scenarios, and only ISS for the 3D scenario. The selection was based on both the RR metric and the computational efficiency demonstrated in Section 4.1. For the 3D keypoint detection methods, we also investigated the performance by applying a uniform subsampling scheme and scoring based on the AUC metric.

*4.2.1 Oakland dataset:* In the first trial, we evaluated the 2D-2D scheme and tested robustness to resolution variation, Gaussian noise and Shot noise using the same parameters described for the evaluation of keypoint detection (Section 4.1.1).

SURF and KAZE were the most robust feature descriptors in response to resolution variation regardless of the keypoint detection method (Fig. 10 (a)). Among the three 2D keypoint detectors we challenged, Fast16-A achieved the highest AUC value and the most robust combination was Fast16-A with the SURF descriptor. In contrast, SIFT, FREAK and BRISK achieved low AUC values at all resolutions regardless of the associated 2D keypoint detector. Interestingly, SIFT, FREAK and BRISK achieved low AUC values even when applied to the original nuisance-free scene.

We also evaluated the robustness of the 2D-2D scheme with various levels of Gaussian noise. SURF and KAZE were again the most robust (Fig. 10 (b)). Similarly to the initial trial, FREAK, BRISK and SIFT achieved low AUC scores regardless of the Gaussian noise level. The same trend was observed in the Shot noise trial (Fig. 10 (c)). In both noise trials, the AUC achieved by each method was quite stable regardless of the noise level, highlighting the robustness of the 2D methods to noise and also the important contribution of the quantization process of Eq. (17).

In the second trial, we considered the 2D-3D scheme. All of the 3D descriptors we tested were sensitive to resolution variation (Fig. 11 (a)), and given the robustness already shown for the 2D keypoint detection methods (Fig. 4 (a)), the low AUC values in the 2D-3D trial were attributed to the 3D descriptors. In contrast, the 2D-3D scheme was more robust against noise nuisances but still inferior to the 2D-2D scheme. For Gaussian and Shot noise (Fig. 11 (b)-(c)), the performance of each 3D descriptor depended strongly on the 2D keypoint detector. Hence, for the GFFT keypoint detector, the best performance was achieved by RoPS, closely followed by HoD-S, FPFH and HoD. A similar trend was apparent for the Fast16-A keypoint detector. However, the FH keypoint detector resulted in higher AUC values for most of the 3D descriptors, with HoD-S and HoD again achieving highest AUC values. Interestingly, TriSI and SHOT achieved a low AUC value regardless of the nuisance applied. Overall, the performance of the 2D-3D scheme was inferior to that of the 2D-2D scheme.

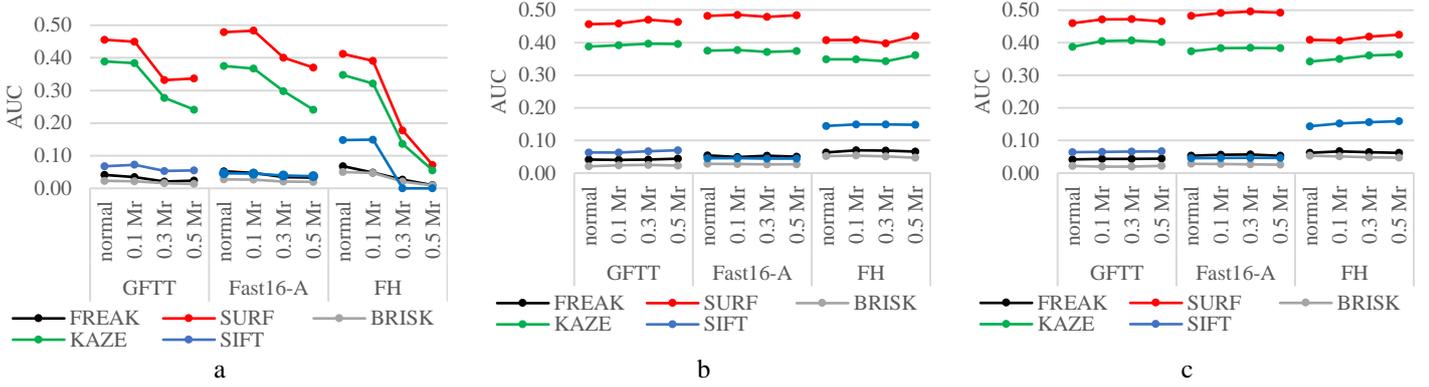

*Fig. 10 Evaluating 2D keypoint detectors and 2D feature descriptors on the Oakland dataset. (a) Resolution variation (b) Gaussian noise (c) Shot noise.*

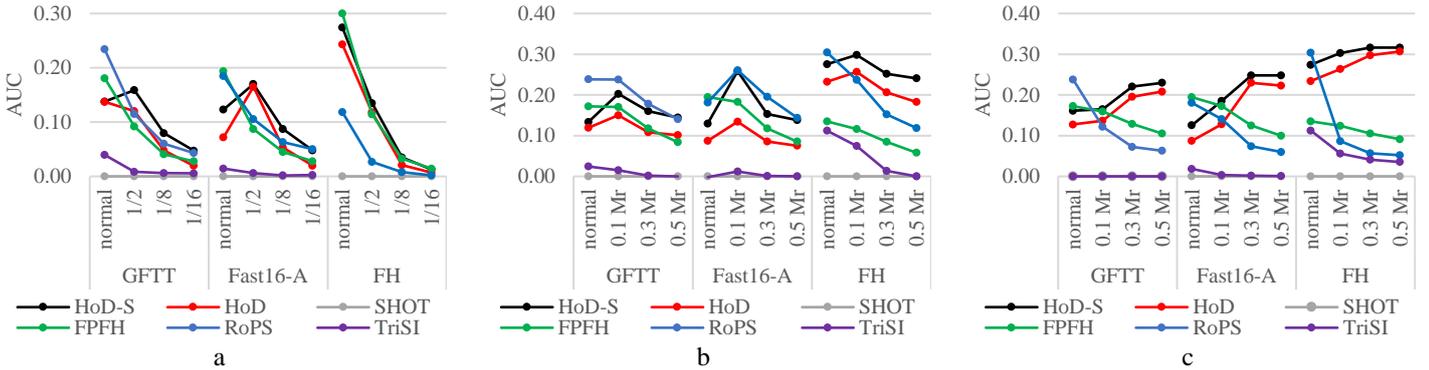

*Fig. 11 Evaluating 2D keypoint detectors and 3D feature descriptors on the Oakland dataset. (a) Resolution variation (b) Gaussian noise (c) Shot noise.*

The third trial was the cross-dimensional 3D-2D scheme. ISS was more robust to resolution variation compared to uniform subsampling (Fig. 12 (a)). Interestingly, the hierarchy between ISS and the uniform subsampling strategy was the same for the 2D-2D and 3D-2D schemes, suggesting that the AUC is mostly affected by the 2D feature descriptors rather than the dimensionality of the keypoint detection method. Even so, the cross-dimensional 3D-2D scheme based on ISS and SURF was more robust to resolution variation, achieving a relatively stable AUC at all resolution levels. The robustness of the 3D-2D scheme to Gaussian noise variation is shown in Fig. 12 (b). The performance of both 3D keypoint detection methods was similar, with ISS gaining a slight advantage. Again, the hierarchy of the 2D-2D and 3D-2D schemes was the same. Finally, we challenged the 3D-2D scheme with various levels of Shot noise (Fig. 12 (c)). SURF achieved the highest performance, although both SURF and KAZE generated appealing AUC scores. Again, the hierarchy of the 2D-2D and 3D-2D schemes was preserved.

The final trial considered the 3D-3D scheme. The performance of this scheme in all three nuisance trials was similar to the 2D-3D scheme, with the 3D-3D scheme showing marginally better AUC values.

Table 4 summarizes the computational time required by each 2D/3D keypint detection and feature description combination, and the average time required by each method regardless of the combination. The most efficient methods were Fast16-A combined with SURF and FH combined with SURF, each requiring only 0.05 s per point cloud.

Given that these combinations involved 2D methods, the processing time not only includes the keypoint detection and feature description methods but also the 3D to multi-2D projection and 2D to 3D back-projection. For the overall performance of the keypoint detection methods considering all description methods, the fastest 2D technique was Fast16-A, and of the 3D methods, uniform subsampling was faster than ISS. For the feature description methods and their overall performance considering all keypoint detection methods, we conclude that most efficient 2D descriptor is SURF, and the most efficient 3D descriptor is HoD-S.

The evaluations on the Oakland dataset lead to the following conclusions:

a. The 2D-2D combination achieves the highest overall performance in terms of AUC and processing efficiency.

b. The 2D feature descriptors preserve their hierarchy and their performance regardless of the keypoint detection dimensionality and method.

c. The 2D feature descriptors are more robust to nuisances than their 3D counterparts. The performance

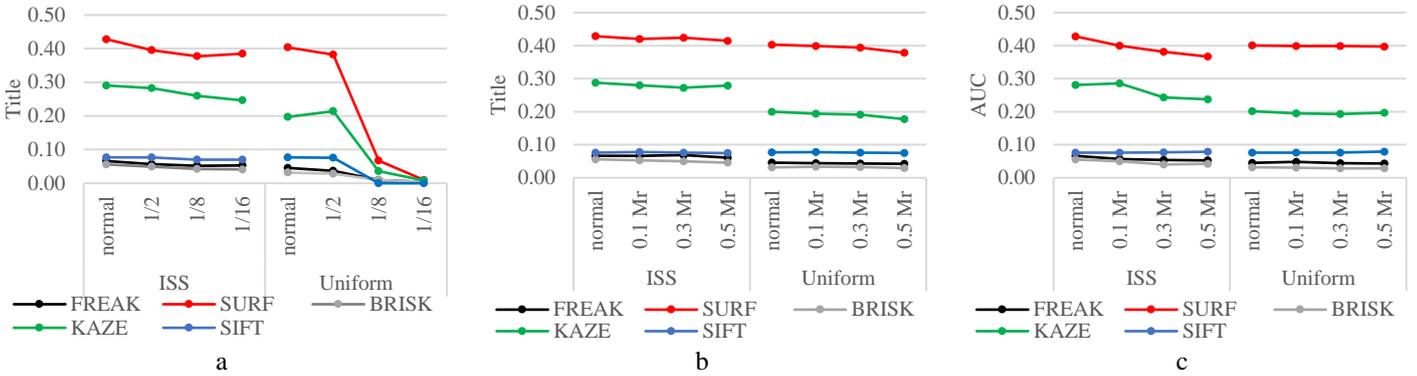

*Fig. 12 Evaluating 3D keypoint detectors and 2D feature descriptors on the Oakland dataset. (a) Resolution variation (b) Gaussian noise (c) Shot noise.*

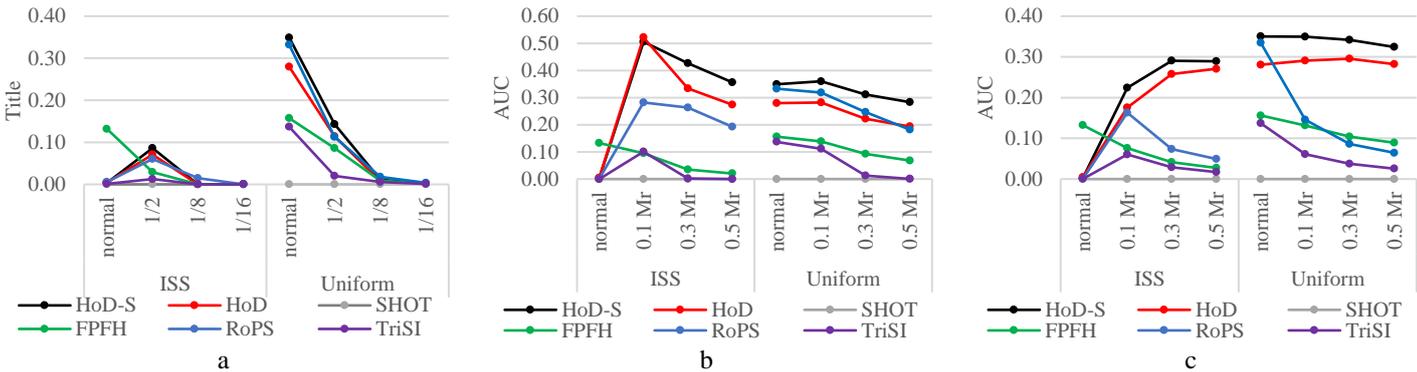

*Fig. 13 Evaluating 3D keypoint detectors and 3D feature descriptors on the Oakland dataset. (a) Resolution variation (b) Gaussian noise (c) Shot noise.*

degradation of the 3D descriptors in response to increasing nuisance levels is also described elsewhere, although in the context of different datasets [4]. Therefore, 3D descriptors appear to generally suffer from low robustness to resolution variation, Gaussian noise and Shot noise.

*4.2.2 Laser Scanner dataset:* As stated above, we only considered the nuisance-free versions of the Laser Scanner, Kinect and SpaceTime datasets. The performance of the various keypoint detection and feature description methods against the Laser Scanner dataset is shown in Fig. 13. For the 2D-2D scheme (Fig. 14 (a)), the highest AUC was achieved by combining SURF with FH, or Fast16-A and SIFT with FH. For the 2D-3D scheme (Fig. 14 (b)), GFTT combined with FPFH performed best, whereas Fast16-A or FH combined with any 3D feature descriptor resulted in the poorest performance. The 3D-2D scheme was the best of all four of the dimensional combinations (Fig. 14 (c)). Specifically, 3D uniform subsampling combined with SURF and KAZE achieved AUC values of 0.77 and 0.73, respectively. This is almost twice the highest value generated by the 2D-2D scheme, three times that of the 2D-3D scheme and five times that of the 3D-3D scheme. Interestingly, the 3D-3D combination, which is the standard approach for 3D data in the form of point clouds, achieved the lowest AUC scores (Fig. 14 (d)) in agreement with earlier studies [4].

*4.2.3 Kinect dataset:* The AUC values representing each combination of methods tested against the standard Kinect dataset are summarized in Fig. 15. The 3D-2D scheme achieved the highest AUC values, specifically ISS combined with SURF.

*4.2.4 SpaceTime dataset:* The AUC values representing each combination of methods tested against the standard SpaceTime dataset are summarized in Fig. 16. Here, the 2D-2D scheme achieved the highest AUC scores, followed by the 3D-2D scheme. Interestingly, however, only SURF and KAZE provided meaningful AUC values.

*4.2.5 Robustness Overall Performance:* To enhance our comparison of the single and cross-dimensional keypoint detection and feature description combinations, the AUC scores achieved by each method averaged over all datasets are presented in Table 5, along with the average performance per keypoint detection and feature description technique on an independent basis.

This analysis shows that the 3D keypoint detectors contribute to higher AUC values, with the 2D FH method following closely behind. Regarding the feature descriptors, SURF clearly achieves the highest AUC values regardless of the keypoint detector, followed by KAZE. RoPS achieves the highest performance among the 3D techniques, but still lower than SURF and KAZE. Overall, the 3D-2D scheme is the most appealing combination, achieving the highest AUC values while imposing among the lowest computational requirements. A detailed analysis of the computational requirements is presented in Table 6.

We also evaluated the performance of each method based on the compactness metric. This reveals the description capability of each feature description technique, but also uses the AUC value so assesses the joint performance of the keypoint detector and feature descriptor. Table 7 presents the average compactness values for all datasets, revealing that ISS and uniform subsampling combined with SURF are the most descriptive combinations.

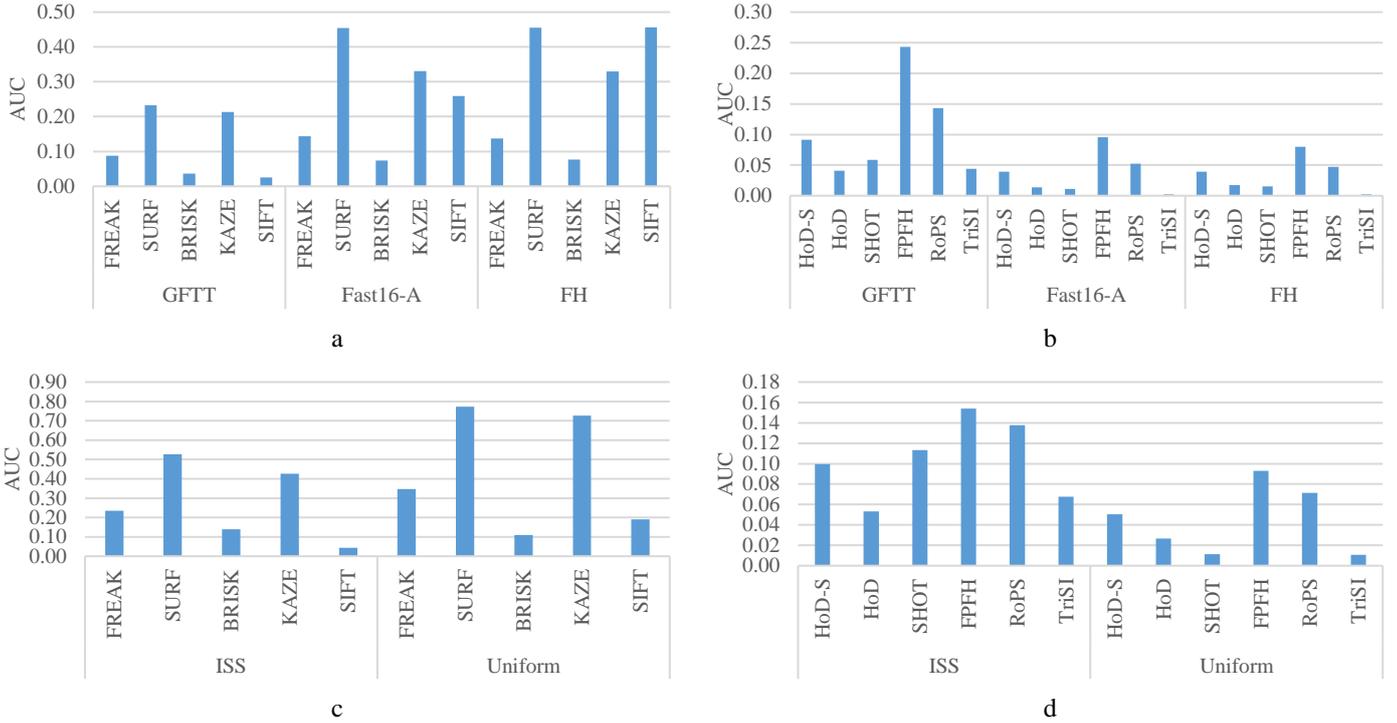

**Fig. 14** *Evaluating keypoint detector and feature descriptor combinations on the Laser Scanner dataset. (a) 2D–2D. (b) 2D–3D. (c) 3D–2D. (d) 3D–3D.*

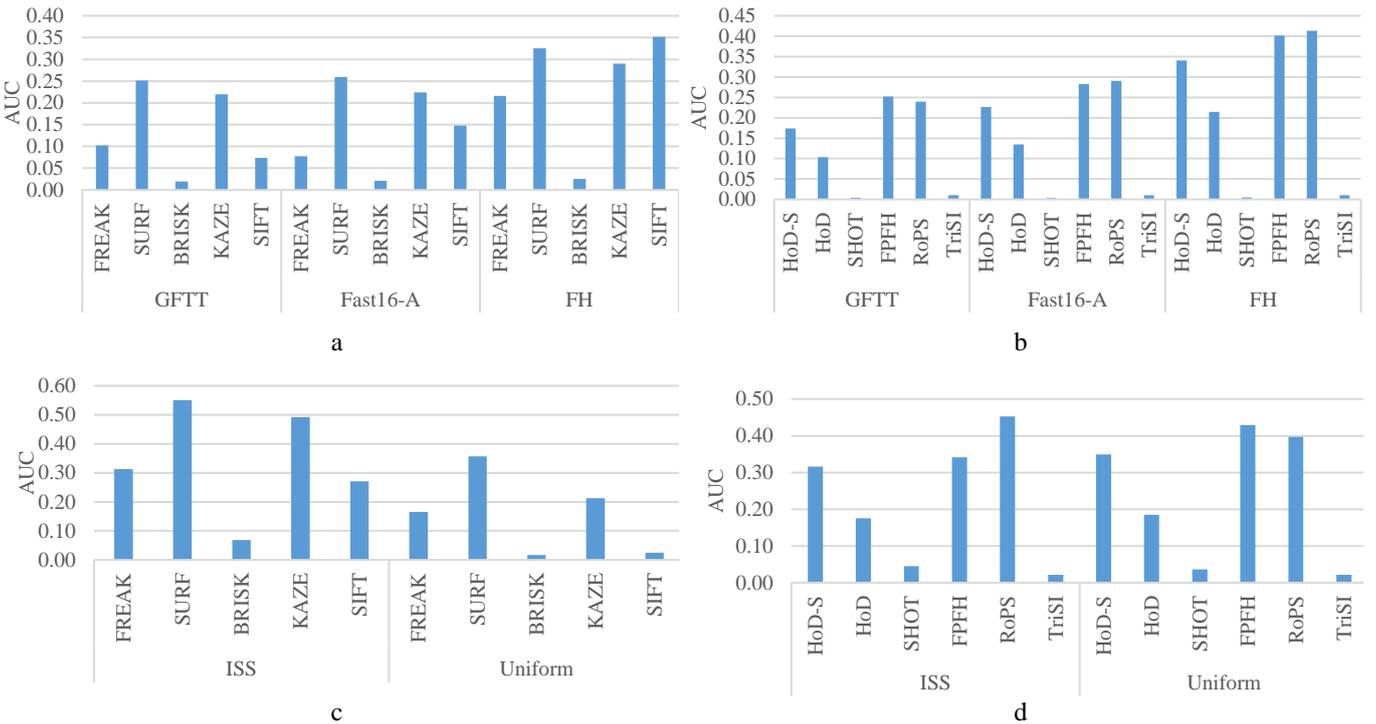

**Fig. 15** *Evaluating keypoint detector and feature descriptor combinations on the Kinect dataset. (a) 2D–2D. (b) 2D–3D. (c) 3D–2D. (d) 3D–3D.*

## 5. Conclusions

In this paper we evaluated single and multi-dimensional combinations of keypoint detection and feature description methods. The descriptiveness of the schemes was evaluated by registration and object recognition on four datasets differing in quality and complexity. The robustness of the schemes to different levels of three nuisance factors (resolution variation, Gaussian and Shot noise) was also examined.

Our evaluation indicated that the optimum dimensionality combination is multi-dimensional,

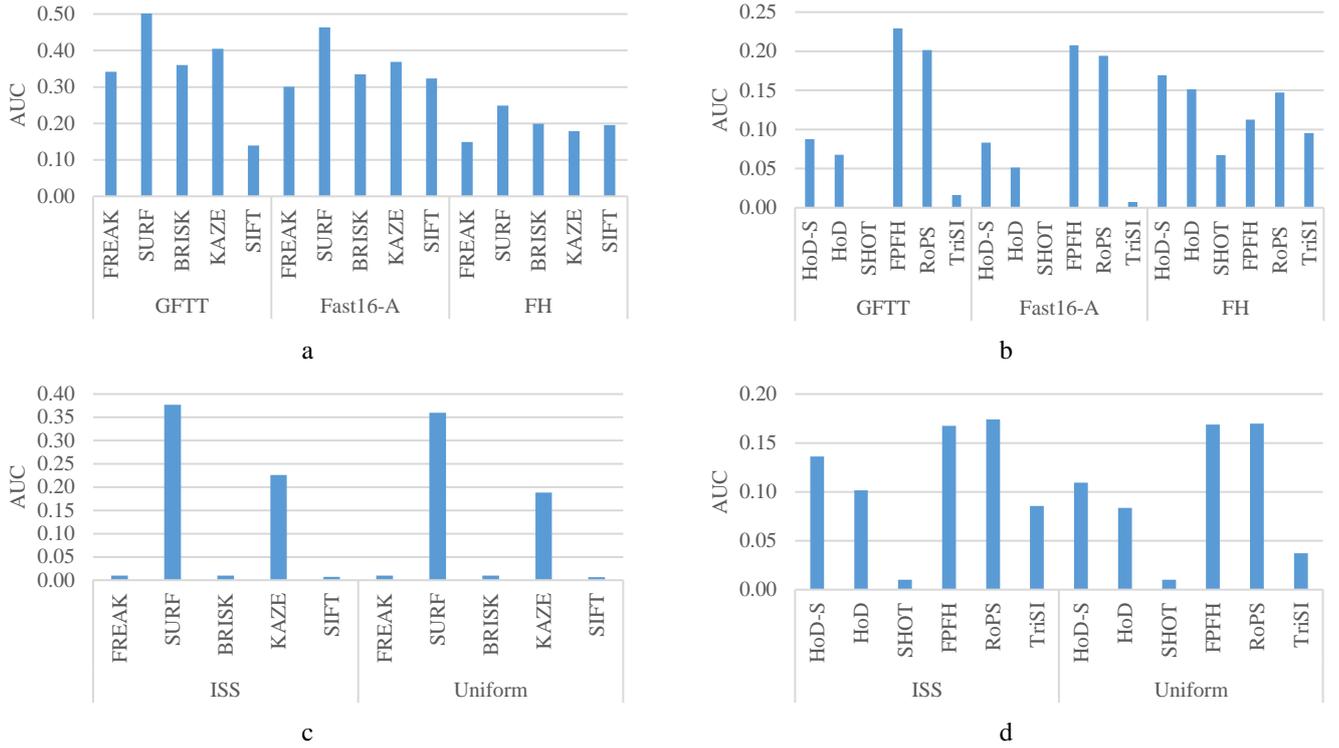

**Fig. 16** *Evaluating keypoint detector and feature descriptor combinations on the SpaceTime dataset. (a) 2D–2D. (b) 2D–3D. (c) 3D–2D. (d) 3D–3D.*

**Table 5** Average AUC Performance on All Datasets

| | | | 2D descriptors | | | | | 3D descriptors | | | | | | |
|---|---|---|---|---|---|---|---|---|---|---|---|---|---|---|
| | | | FREAK | SURF | BRISK | KAZE | SIFT | HoD-S | HoD | SHOT | FPFH | RoPS | TriSI | average |
| Keypoint detectors | 2D | GFTT | 0.14 | 0.36 | 0.11 | 0.31 | 0.08 | 0.12 | 0.09 | 0.02 | 0.23 | 0.20 | 0.03 | 0.15 |
| | | Fast16-A | 0.14 | 0.41 | 0.11 | 0.32 | 0.20 | 0.12 | 0.09 | 0.01 | 0.20 | 0.18 | 0.01 | 0.16 |
| | | FH | 0.15 | 0.36 | 0.09 | 0.29 | 0.29 | 0.21 | 0.16 | 0.03 | 0.18 | 0.23 | 0.06 | 0.18 |
| | 3D | ISS | 0.16 | 0.47 | 0.07 | 0.36 | 0.10 | 0.14 | 0.08 | 0.04 | 0.20 | 0.26 | 0.07 | 0.18 |
| | | Uniform | 0.14 | 0.47 | 0.04 | 0.33 | 0.08 | 0.22 | 0.14 | 0.02 | 0.21 | 0.24 | 0.05 | 0.18 |
| | | average | 0.15 | 0.42 | 0.09 | 0.32 | 0.15 | 0.16 | 0.11 | 0.02 | 0.20 | 0.22 | 0.04 | |

**Table 6** Total Processing time

| | | | 2D descriptors | | | | | 3D descriptors | | | | | | |
|---|---|---|---|---|---|---|---|---|---|---|---|---|---|---|
| | | | FREAK | SURF | BRISK | KAZE | SIFT | HoD-S | HoD | SHOT | FPFH | RoPS | TriSI | average |
| Keypoint detectors | 2D | GFTT (F) | 0.23 | 0.11 | 0.92 | 0.14 | 0.26 | 2.64 | 4.05 | 8.20 | 54.78 | 26.30 | 44.92 | 12.96 |
| | | Fast16 (A) | 0.16 | 0.05 | 0.80 | 0.07 | 0.26 | 0.83 | 1.17 | 5.19 | 26.51 | 7.69 | 29.61 | 6.58 |
| | | FH (A) | 0.61 | 0.05 | 0.80 | 0.07 | 0.05 | 0.65 | 0.93 | 4.82 | 34.34 | 5.95 | 29.61 | 7.08 |
| | 3D | ISS | 0.29 | 0.20 | 0.94 | 0.29 | 0.26 | 1.84 | 2.74 | 6.54 | 59.46 | 21.39 | 46.35 | 12.75 |
| | | Uniform | 0.20 | 0.11 | 0.84 | 0.15 | 0.27 | 0.98 | 1.40 | 5.19 | 53.23 | 9.43 | 35.15 | 9.72 |
| | | average | 0.30 | 0.10 | 0.86 | 0.14 | 0.22 | 1.39 | 2.06 | 5.99 | 45.66 | 14.15 | 37.13 | |

**Table 7** Average Compactness on All Datasets

| | | | 2D descriptors | | | | 3D descriptors | | | | | | average |
|---|---|---|---|---|---|---|---|---|---|---|---|---|---|
| | | | FREAK | SURF | BRISK | KAZE | SIFT | HoD-S | HoD | SHOT | FPFH | RoPS | TriSI | |
| Keypoint detectors | 2D | GFTT | 2.23 | 5.63 | 1.72 | 4.77 | 0.61 | 3.06 | 0.36 | 0.05 | 6.82 | 1.50 | 0.04 | 2.43 |
| | | Fast16-A | 2.23 | 6.45 | 1.76 | 5.04 | 1.52 | 2.94 | 0.38 | 0.01 | 5.91 | 1.31 | 0.01 | 2.51 |
| | | FH | 2.27 | 5.66 | 1.41 | 4.49 | 2.27 | 5.13 | 0.65 | 0.07 | 5.53 | 1.69 | 0.09 | 2.66 |
| | 3D | ISS | 2.42 | 7.38 | 1.09 | 5.63 | 0.78 | 3.50 | 0.34 | 0.11 | 5.98 | 1.94 | 0.10 | 2.66 |
| | | Uniform | 2.23 | 7.38 | 0.66 | 5.20 | 0.61 | 5.38 | 0.59 | 0.04 | 6.44 | 1.80 | 0.08 | 2.76 |
| | | average | 2.27 | 6.50 | 1.33 | 5.02 | 1.16 | 4.00 | 0.46 | 0.06 | 6.14 | 1.65 | 0.07 | |

contrasting with the typical approach for keypoint detection and feature description in 3D data based on methods specifically designed for the 3D data domain. We found that the optimum combination is a cross-dimensional scheme combining the ISS/uniform subsampling 3D keypoint detection method with the SURF 2D feature description method. These combinations achieve the highest descriptiveness with a very low processing time, combining the advantages of 2D and 3D processes.